\documentclass[nohyperref]{article}

\usepackage{microtype}
\usepackage{graphicx}
\usepackage{subfigure}
\usepackage{booktabs} % for professional tables

\usepackage{hyperref}

\usepackage[accepted]{icml2023}

\usepackage{amsmath}
\usepackage{amssymb}
\usepackage{mathtools}
\usepackage{amsthm}

\usepackage[capitalize,noabbrev]{cleveref}

\theoremstyle{plain}

\theoremstyle{definition}

\theoremstyle{remark}

\usepackage[textsize=tiny]{todonotes}
\usepackage{bbm}

\DeclareMathOperator*{\argmax}{arg\,max}

\begin{document}

\twocolumn[
% \icmltitle{Submission and Formatting Instructions for \\
           % International Conference on Machine Learning (ICML 2023)}
% \icmltitle{Retrosynthetic Planning with Tree-Structured Reinforcement Learning}
%\icmltitle{Planning with Dual Value Network for Retrosynthetic Planning}
\icmltitle{Retrosynthetic Planning with Dual Value Networks}
% It is OKAY to include author information, even for blind
% submissions: the style file will automatically remove it for you
% unless you've provided the [accepted] option to the icml2023
% package.

% List of affiliations: The first argument should be a (short)
% identifier you will use later to specify author affiliations
% Academic affiliations should list Department, University, City, Region, Country
% Industry affiliations should list Company, City, Region, Country

% You can specify symbols, otherwise they are numbered in order.
% Ideally, you should not use this facility. Affiliations will be numbered
% in order of appearance and this is the preferred way.
\icmlsetsymbol{equal}{*}

\begin{icmlauthorlist}
\icmlauthor{Guoqing Liu}{equal,msr}
\icmlauthor{Di Xue}{equal,nju}
\icmlauthor{Shufang Xie}{ruc}
\icmlauthor{Yingce Xia}{msr}
\icmlauthor{Austin Tripp}{cu}
\\
\icmlauthor{Krzysztof Maziarz}{msr}
\icmlauthor{Marwin Segler}{msr}
\icmlauthor{Tao Qin}{msr}
\icmlauthor{Zongzhang Zhang}{nju}
\icmlauthor{Tie-Yan Liu}{msr}
%\icmlauthor{}{sch}
%\icmlauthor{}{sch}
\end{icmlauthorlist}

\icmlaffiliation{nju}{National Key Laboratory for Novel Software Technology, Nanjing University}
\icmlaffiliation{cu}{University of Cambridge}
\icmlaffiliation{msr}{Microsoft Research AI4Science}
\icmlaffiliation{ruc}{Renmin University of China}

\icmlcorrespondingauthor{Tao Qin}{taoqin@microsoft.com}
% \icmlcorrespondingauthor{Firstname2 Lastname2}{first2.last2@www.uk}

% You may provide any keywords that you
% find helpful for describing your paper; these are used to populate
% the "keywords" metadata in the PDF but will not be shown in the document
\icmlkeywords{Machine Learning, ICML}

\vskip 0.3in
]

% this must go after the closing bracket ] following \twocolumn[ ...

% This command actually creates the footnote in the first column
% listing the affiliations and the copyright notice.
% The command takes one argument, which is text to display at the start of the footnote.
% The \icmlEqualContribution command is standard text for equal contribution.
% Remove it (just {}) if you do not need this facility.

\printAffiliationsAndNotice{\icmlEqualContribution}  % leave blank if no need to mention equal contribution

\begin{abstract}
% Abstracts must be a single paragraph, ideally between 4--6 sentences long.
% Gross violations will trigger corrections at the camera-ready phase.

Retrosynthesis, which aims to find a route to synthesize a target molecule from commercially available starting materials, is a critical task in drug discovery and materials design.  
% has drawn much attention in the machine learning (ML) community recently. 
% ML-based retrosynthesis relies on a single-step predictor to split a molecule into reactants and a multi-step planner to search for a complete synthesis route by iteratively calling the predictor. 
Recently, the combination of ML-based single-step reaction predictors with multi-step planners has led to promising results. 
% In most ML-based methods today, the single-step prediction model is trained to optimize the single-step retrosynthesis accuracy, without considering complete retrosynthesis routes. 
However, the single-step predictors are mostly trained offline to optimize the single-step accuracy, without considering complete routes.  
% In this work, we propose to enhance the single-step model by optimizing both the single-step accuracy and complete routes. 
% In this work, we propose to enhance the single-step model by optimizing the complete routes while retaining the single-step accuracy.
% In particular, we leverage reinforcement learning (RL) and self-generated experiences to improve the single-step model. 
% In this work, we leverage reinforcement learning (RL) and self-generated experiences to enhance the single-step model. 
Here, we leverage reinforcement learning (RL) to improve the single-step predictor, by using a tree-shaped MDP to optimize 
% the likelihood of 
complete routes. 
% while retaining single-step accuracy. 
% [to add more details here] 
%Specifically, we formulate the retrosynthesis problem with a Retrosynthesis MDP. 
% where the trajectory of each episode is tree-structured. 
% To leverage 
% Motivated by 
% the planning nature and decomposability of retrosynthesis,
% Motivated by 1) retrosynthesis is a planning problem; 2) the goal of retrosynthesis can be decomposed into two smaller subgoals: successful and low-cost. 
% The goal of retrosynthesis is to find routes that are synthesizable and as low-cost as possible.
% Motivated by this, 
% Motivated by the fact that retrosynthesis usually requires routes that are synthesizable and as low-cost as possible.
% The goal of retrosynthesis can be naturally decomposed into 1) synthesizable; 2) as low-cost as possible. 
Specifically, we propose a novel online training algorithm, called Planning with Dual Value Networks (PDVN), which alternates between the planning phase and updating phase. 
In PDVN, we construct two separate value networks to predict the synthesizability and cost of molecules, respectively.
% Since desirable routes should be both synthesizable and of low cost, we construct two separate value networks to predict the synthesizability and cost of molecules, respectively.
To maintain the single-step accuracy, 
% for
% predicting valid reactions, 
% we design a two-branch policy network structure.
we design a two-branch network structure for the single-step predictor.
% We design a realistic policy structure to maintain single-step accuracy.
% make the synthetic route more realistic.  
% to generate both realistic and successful routes.
% To improve the training efficiency, we also introduce an extra-lesson training strategy. 
% Results on the widely used USPTO dataset show the potential of our method, in terms of both the plan efficiency and route quality (\textcolor{black}{\textbf{Show number:} 
% On the widely used USPTO benchmark, our method significantly improves the search success rate of both  Retro* planner and Retrograph planner from xxx to xxx, from xxx to xxx. Reduce length from xxx to xxx.
% \color{}{
On the widely-used USPTO dataset, our PDVN algorithm improves the search success rate 
% of addressing the retrosynthesis problem 
of existing multi-step planners (e.g., increasing the success rate from $85.79\%$ to $98.95\%$ for Retro*).
% and reducing the number of model calls by half while solving $99.47\%$ molecules for RetroGraph).
% achieving state-of-the-art on the USPTO dataset
% ). 
Additionally, PDVN helps find shorter synthesis routes
% reduces the length of the found synthesis routes 
(e.g., reducing the average route length from $5.76$ to $4.83$ for Retro*).
% and from $5.63$ to $4.78$ for RetroGraph)
Our code is available at \url{https://github.com/DiXue98/PDVN}.
% while reducing the length of the synthesis route 
% Besides, we also found that our method can bring performance gains in the harder molecules from ChEMBL dataset and GDB17 dataset (e.g, in ChEMBL-1000, the number of molecules solved is increased from 762 to 846, and in GDB17-1000, the number of molecules solved is increased from 95 to 252). 
% }

% and improve the search success rate of the Retrograph planner from xxx to xxx. 
% to 99.47%, advancing previous state-of-the-art performance for
% 2.6 points. ). 
\end{abstract}

\section{Introduction}

\begin{figure}[t]
% \centerline{\includegraphics[width=0.4\textwidth]{illustration_of_retrosynthesis_planning.png}}
\centerline{\includegraphics[width=0.5\textwidth]{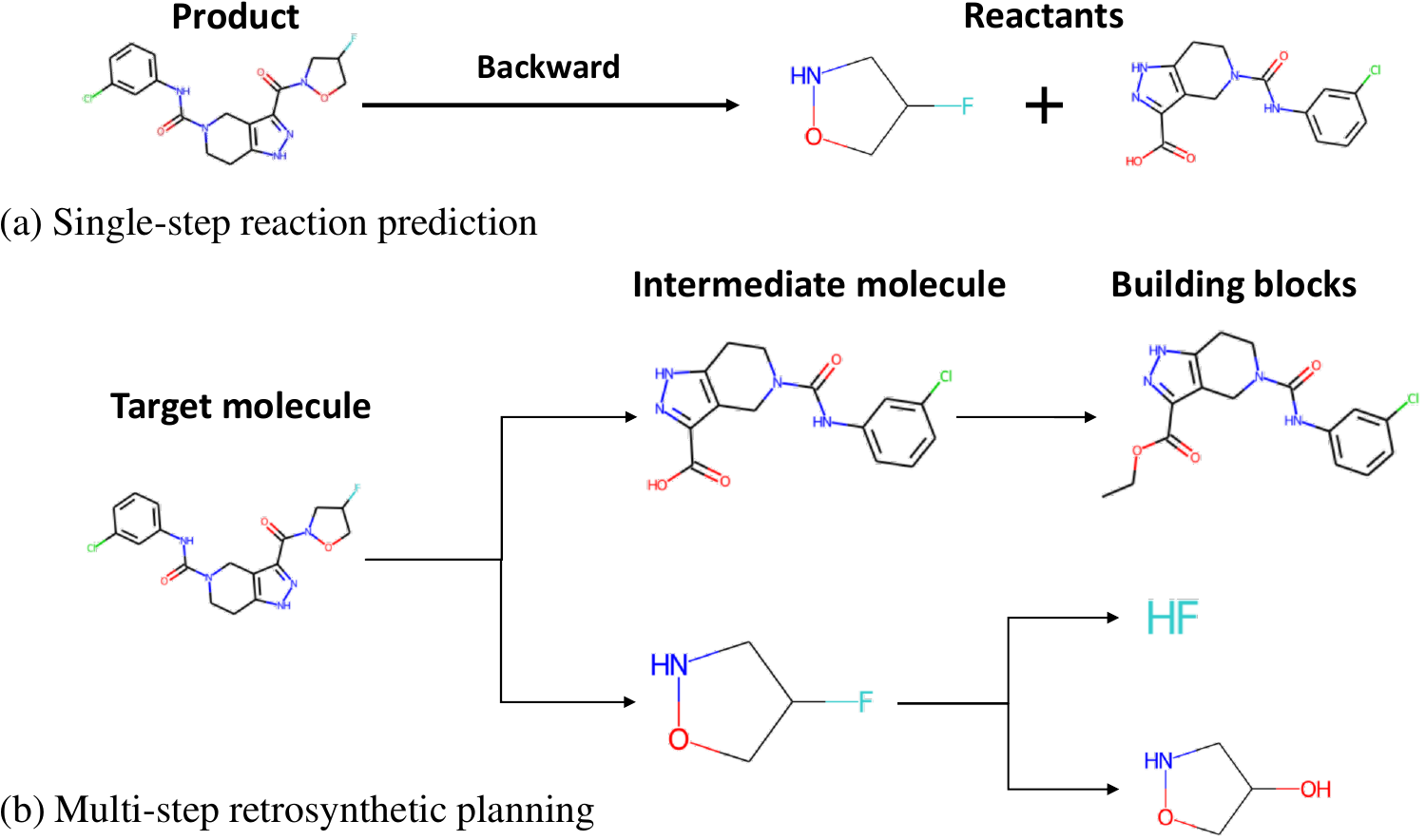}}
\caption{
% {Illustration of retrosynthesis: 
a) The single-step reaction predictor, which predicts potential ways to break a molecule into reactants at each step. 
b) The multi-step planner, which searches for a complete route by iteratively calling the predictor.
The goal of retrosynthesis is to find a synthesis route ending up
in the building block molecules for a target molecule. 
}
\label{fig:retrosynthesis}
\end{figure}

% \textbf{Background of the retrosynthesis problem: 1) impact; 2) what is retro?}
% Discovering reaction pathways to synthesize a target molecule using a set of available building block molecules 
% % using a set of known or commercially available building block molecules 
% is a fundamental problem in organic chemistry, 
% widely used in important real-world applications such as drug discovery (cite) and material design (cite).
% % Retrosynthesis planning aims at finding a series of chemically valid reactions starting
% % from the target molecule until reaching the commercially available building block
% % molecules in a backward and recursive manner.
% Retrosynthesis planning aims at identifying a series of reactions that can lead to the synthesis of the product, by searching backwards and
% iteratively applying chemical transformations to unavailable
% molecules. As thousands of theoretically-possible transformations can all be applied during each step of reactions,
% the search space of planning will be huge and makes the
% problem challenging even for experienced chemists.

% 1/7/2022

\begin{figure*}[t]
\centerline{\includegraphics[width=0.9\textwidth]{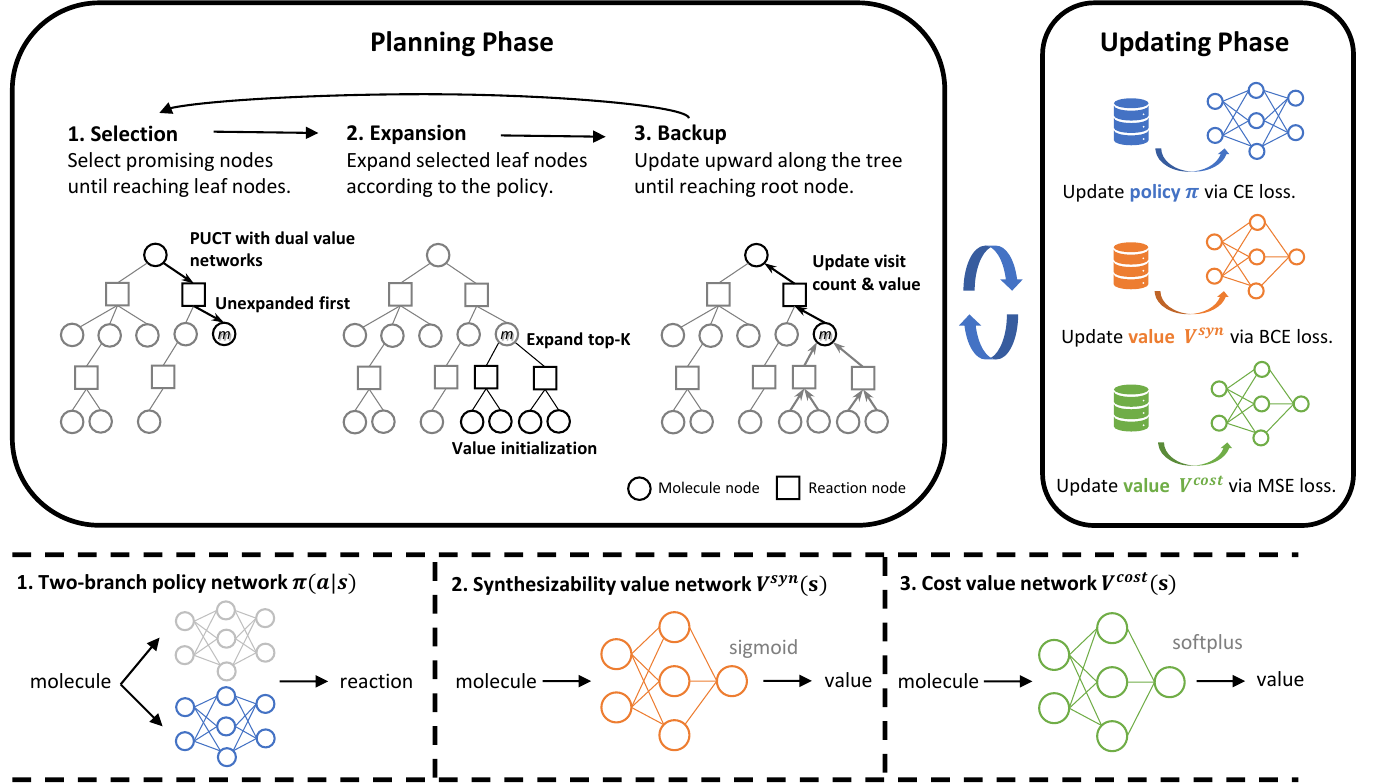}}
\caption{
An illustration of our PDVN algorithm. 
The PDVN algorithm has three modules: 
1) a two-branch policy network;
2) a synthesizability value network that predicts if a molecule can be synthesized;
3) a cost value network that predicts the required synthesis cost if synthesizable.
% which includes two steps and three modules.  
PDVN is initialized with an offline SL model and alternates between two phases: 
1) \textit{Planning phase}: 
% generate the simulated synthesis experiences on the tree-shaped MDP with the help of the policy network and dual value networks. 
simulate synthesis experiences on the tree-shaped MDP under the guidance of the policy network and dual value networks. 
2) \textit{Updating phase}: extract useful training targets from the generated experiences and update all three networks.
Finally, the single-step model trained by PDVN is plugged into popular multi-step planners to enhance their performance. 
}
\label{fig:framework}
\end{figure*}

Retrosynthesis is one of the fundamental problems in organic chemistry, widely used in important applications such as drug discovery and materials design.
Given a target molecule, the goal of retrosynthesis is to identify a series of chemically valid reactions  starting from the target molecule until reaching commercially available building block molecules in a backward and recursive manner.
There are many theoretically-possible transformations that can be applied 
% during each step of reactions. 
at each step. 
In addition, each intermediate molecule could be broken into several reactants in one reaction.  
As a result, the search space of retrosynthesis is enormous and makes the problem challenging even for experienced chemists.

Retrosynthesis has drawn much attention in the machine learning (ML) community in recent years~\cite{Segler2018PlanningCS, dl4retro_survey}. 
As shown in Fig.~\ref{fig:retrosynthesis}, current ML-based retrosynthesis consists of 
1) a single-step reaction predictor to predict a set of potential reactions given a molecule; 
2) a multi-step planner to search for a complete synthesis route by iteratively calling the predictor. 
% The single-step reaction prediction, which predicts a set of possible direct reactants given molecule, serves as the foundation for realizing the multi-step retrosynthesis. 
% Recently,  
Many algorithms have been proposed for the single-step predictor through
% For the single-step model, there have been many approaches that train deep neural networks by 
supervised learning (SL) based on existing real-world reaction datasets~\cite{lowe2012extraction}, such as template-based methods~\cite{segler2017neural, coley2017retrosim, dai2019retrosynthesis} and template-free methods~\cite{liu2017s2s, tetko2020state}. 
% Using the existing real-world reaction datasets~\cite{Lowe2012ExtractionOC}, recent methods train deep neural networks in a supervised manner to predict a reactant-set from a given product.  
% Existing work roughly fall into two categories, either template-based or template-free.  
% Existing methods include the template-based ones~\cite{segler2017neural, coley2017retrosim, dai2019retrosynthesis, shuan2021localretro}, and the template-free ones~\cite{liu2017s2s, tetko2020state}. 
% zheng2020selfcorrected}. 
% On the other hand, 
Researchers have also developed several search algorithms for multi-step planning, such as 3N-MCTS~\cite{Segler2018PlanningCS}, Retro*~\cite{chen20retrostar}, and RetroGraph~\cite{xie2022retrograph}.

% search algorithms for retrosynthetic planning based on the
% DNN-based single-step retrosynthetic models. Their main
% idea is to represent retrosynthetic planning as a sequential
% decision making problem and apply tree search algorithms
% such as Monte Carlo tree search (Segler et al., 2018), proof
% number search (Kishimoto et al., 2019), and A* search
% (Chen et al., 2020b)

% \textbf{How does current work address the problem? and point out the problem of those work.}
% The main challenge of retrosynthesis planning is twofold: (a) finding
% an accurate single-step retrosynthetic model that predicts
% a single reaction of a given product and (b) designing an
% efficient search algorithm for a reaction pathway starting
% from the set of building block molecules.

% \begin{figure*}[t]
% \centerline{\includegraphics[width=0.9\textwidth]{R-MDP.png}}
% % \caption{R-MDP. 
% % Given a target molecule $m_{0}$, recursively expand the tree by executing the policy $\pi(r|m)$, until the buyable molecules or dead molecules are encountered.
% % } 
% \caption{(Placeholder) \textcolor{black}{Illustration of our framework}, which includes the two steps and three modules.  }
% \label{fig:R-MDP}
% \end{figure*}

% \textbf{Contributions of our work: method and experiment.} 

However, in most ML-based methods today, the single-step predictor is usually trained offline to optimize the single-step accuracy, without considering complete synthesis routes. 
In this work, we leverage reinforcement learning (RL)  
to improve the single-step predictor, 
or the policy network in the terminology of RL~\cite{sutton2018reinforcement, dac2021},
to optimize complete routes.
% Specifically, 
% we first formulate the retrosynthesis problem as a Retrosynthesis MDP.
% Specifically, we first use a tree-shaped MDP to formulate the retrosynthesis problem.
% Since a product molecule can be broken into several reactants at each step, the trajectory of each episode is tree-shaped in such an MDP. 
% Since a product molecule can be broken into several reactants at each step, we formulate the retrosynthesis problem using a tree-shaped MDP. 
To do this, we use a tree-shaped Markov Decision Process (MDP) to formulate the retrosynthesis problem.
% the trajectory of each episode is tree-shaped in such an MDP., and thus  
% Specifically, wwe first use a tree-shaped MDP to formulate s a tree-shaped MDP.
% Then, motivated by the planning nature and the decomposability of the retrosynthesis problem,
Then, we propose a novel online training algorithm, called \textit{Planning with Dual Value Networks (PDVN)}, which alternates between two phases:
% In the Planning phase, we simulate the synthesis experiences on the tree-shaped MDP under the guidance of the policy network and dual value networks. 
% In the Updating phase, we carefully extract useful training targets from the generated experiences to update the policy network and dual value networks. 
\begin{enumerate}
\item \textit{Planning phase:} Given a batch of training target molecules, 
we generate the simulated experiences by planning with the dual value networks.
\item \textit{Updating phase:} We carefully extract useful training targets from the generated experiences, and update the policy network and dual value networks.
\end{enumerate}

Since desirable routes in retrosynthesis should be both synthesizable and low-cost, we construct two separate networks 
% i.e., the dual value networks, 
to predict the synthesizability and synthesis cost of molecules.
% 1) Planning phase: given a batch of training target molecules, we generate the simulated experiences by planning with the dual value networks and policy network; 2) Updating phase: we extract useful training targets from the generated experiences, and update the policy network and dual value networks.  
% Motivated by the fact that the desirable synthesis routes should be 1) synthesizable and 2) as low-cost as possible,
% Since desirable routes in retrosynthesis should be both synthesizable and low-cost, we construct two separate value networks to predict the synthesizability and synthesis cost of molecules.
% During the planning phase of PDVN, the dual value network will lead to a new selection rule.
% During the updating phase of PDVN, we update each value network in its own way.
% Our PDVN algorithm 
% according to its own definition. 
To retain the single-step accuracy, we design a two-branch policy network structure. 
% One branch is a fixed single-step model that is pre-trained by SL to provide a set of valid reactions~(e.g., top $50$ candidates).
% The other branch is a learnable single-step model that optimizes the probability distribution over valid reactions to optimize complete routes.
The first branch is a fixed, pre-trained single-step model that provides a set of valid reactions (e.g., the top 50 candidates). The second branch is a learnable single-step model that optimizes the probability distribution over valid reactions to optimize complete routes.

% to optimize the probability distribution over valid reactions towards the goal of retrosynthetic planning.
% synthesizability and low cost. 
% given the provided set. 

To demonstrate the effectiveness of our PDVN algorithm, we conduct extensive experiments on the widely used USPTO dataset~\cite{lowe2012extraction, chen20retrostar}. 
The results show that using the single-step model trained by PDVN largely improves the success rate and route quality of existing multi-step planners.
% our PDVN algorithm can 
% compared with
% when added to the
% existing multi-step planners. 
For the Retro* planner~\cite{chen20retrostar}, PDVN increases the search success rate from $85.79\%$ to $98.95\%$.
% under the limit of 500 model calls;
% Meanwhile, 
% Remarkably, PDVN+Retro* solves more test molecules (e.g., 90.53\% v.s. 85.79\%) using only 50 model calls, while Retro* solves 85.79\%   .
For the RetroGraph planner~\cite{xie2022retrograph}, PDVN reduces the number of model calls by half when solving $99.47\%$ molecules, 
% increases the success rate to $100\%$, 
and achieves the state-of-the-art on the USPTO dataset.
% Besides, PDVN also effectively reduce the length of 
% Thanks to imitating reactions that are realistic and
% executable from building block molecules, our framework
% significantly improves the success rate of solving the retrosynthetic problem from 86.84\% to 96.32\%. 
% Moreover, the
% reduced average time for planning demonstrates the efficacy
% of our framework. 
Additionally, PDVN effectively 
reduces the length of found synthesis routes. 
We also find that PDVN can bring performance gains when addressing hard target molecules from ChEMBL and GDB17 datasets. 
For the ChEMBL-1000 dataset, the number of molecules solved increased from $762$ to $835$.
For the GDB17-1000 dataset, the number of molecules solved increased from $95$ to $269$. 
Case studies show that PDVN can reliably find chemically valid synthesis routes.  
% In our ablation studies, we
% show the effectiveness of each component in our framework.

% Our codes are available at \url{https://github.com/DiXue98/PDVN}.

% \section{Preliminary}
\section{Related Work}

% \subsection{Single-Step retrosynthesis}
\paragraph{Single-Step Retrosynthesis.}
Denote the space of all molecules as $\mathcal{S}$. 
The single-step predictor takes a product molecule $s \in \mathcal{S}$ as input and predicts a set of possible reactants that can be used to synthesize $s$.
A single-step model can be learned from existing real-world datasets of chemical reactions.  
% The existing s
Current single-step models roughly fall into two categories, i.e., template-based and template-free.
Template-based methods predict reactants with reaction templates that encode chemical reaction cores. 
The key is to rank template candidates and select an appropriate one to apply.
Recent works~\cite{segler2017neural, coley2017retrosim, dai2019retrosynthesis, shuan2021localretro} address the problem of template selection by using a classification neural network.
On the other hand, inspired
by the recent progress of seq2seq models~\cite{sutskever2014sequence} and Transformers~\cite{vaswani2017attention},
template-free methods~\cite{liu2017s2s, tetko2020state} cast single-step retrosynthesis as a translation task, where SMILES string\footnote{Symbolic representation for describing the structures of molecules using ASCII strings.} of a product molecule is translated to these of the reactants.
% As more single-step models develop, we observed increasing accuracy in the predictions of molecular disconnections. 
% mention those benchmark papers, such as Mind the Retrosynthesis Gap: Bridging the divide between Single-step and Multi-step Retrosynthesis Prediction. Echo our motivation. 
As more single-step models are developed, the single-step accuracy continues to increase. 
Some recent benchmark papers~\cite{hassen2022mind, tu2022retrosynthesis} show that the single-step models need to be developed and tested for the multi-step domain.

% \subsection{Multi-step retrosynthesis}
\paragraph{Multi-Step Retrosynthesis.}
% Instead of predicting the single-step reaction, 
Multi-step retrosynthetic planning aims to search for the whole synthesis route, by iteratively calling the single-step model. 
\cite{segler2017towards, Segler2018PlanningCS} used a Monte Carlo Tree Search (MCTS) algorithm to plan the synthesis routes of small organic molecules. 
\cite{akihiro2019dfpn} propose a DFPN-E method that combines Depth-First Proof-Number Search~(DFPN) with heuristic edge initialization.   
Recently, \cite{chen20retrostar} propose Retro*, a neural-based A*-like algorithm, which employs AND-OR search trees and adopts the best-first search strategy on the AND-OR tree.
To reduce the duplication of molecules in the tree-based search method, \cite{xie2022retrograph} propose a graph-based search algorithm named RetroGraph, to further improve the performance of A*-like search algorithms. 
 
\cite{kim2021self} propose a self-improving procedure, called Retro*+, which trains a single-step model to imitate successful trajectories found by itself. 
Despite its achievements, Retro*+ only maximizes the success rate and leverages successful simulated experiences to improve the single-step model. 
Additionally, the A*-like search algorithm used in Retro*+ is based on the best-first search and may fail to effectively balance exploration and exploitation when generating experiences. 
\cite{yu2022grasp} propose GRASP, a goal-driven actor-critic method for finding routes with a specific prescribed goal such as building block materials. 
% Different from 
Unlike GRASP, which
% GRASP focusing 
focuses on goal-driven retrosynthesis, our work focuses more on general retrosynthetic planning.

% Given a target molecule, multi-step retrosynthesis aims at searching for a successful and low-cost synthesis route.
% The relation between our and recent work, like retro*+, EG-MCTS, GRASP. 

\section{Method}
% 1/4/2022
% In this section, we first propose the tree-structured MDP for retrosynthesis on which we can optimize the cost of the synthesizing routes in \ref{sec:formulation}. Then we introduce how to plan for successful and low-cost routes using the MCTS technique in \ref{sec:planning}. To improve the models with the data collected during planning, we detail the data collection protocol and the training targets. In \ref{sec:realistic}, we show how to avoid unrealistic reactions by using the pretrained model smartly.

% 1/9/2022
% \textcolor{black}{In this section, we first formulate the retrosynthesis problem into the retrosynthesis MDP, where the trajectory is tree-structured (section~\ref{sec:formulation}). Then, motivated by the nature of the problem formulation and the fact that retrosynthesis aims to obtain routes that are synthesizable and low-cost, we propose a novel planning algorithm (section~\ref{sec:planning}) that unifies these two objectives to improve the one-step model (section~\ref{sec:training}). To retain one-step accuracy, we propose a realistic policy structure for the one-step model (section~\ref{sec:realistic}).}

\textcolor{black}{In this section, we first formulate the retrosynthesis problem using a tree-shaped MDP~(Section~\ref{sec:formulation}). Then, 
% motivated by the fact that the goal of retrosynthesis is to obtain routes that are 1) synthesizable; 2) as low-cost as possible, 
we introduce the Planning with Dual Value Networks (PDVN) algorithm, which alternately performs the planning phase (Section~\ref{sec:planning})
and the updating phase (Section~\ref{sec:training}). 
Finally, to retain single-step retrosynthesis accuracy, we introduce a two-branch policy network structure~(Section~\ref{sec:realistic}).}
% The overview of our PDVN algorithm can refer to Fig.~\ref{fig:framework}. 
For an overview of our PDVN algorithm, refer to Fig.~\ref{fig:framework}.

\subsection{\textcolor{black}{Retrosynthesis MDP}}\label{sec:formulation}

% The ultimate goal of retrosynthesis is to search for feasible synthesizing routes while keeping the overall cost as little as possible. Past work mainly focuses on the synthesizability of the target molecules and leaves the 
% To encode this requirement into the problem, we propose a novel tree-structured MDP to capture the costs of the synthetic routes.

% \textbf{\#figure illustrating MDP}. 

\begin{figure}[t]
\centerline{\includegraphics[width=0.5\textwidth]{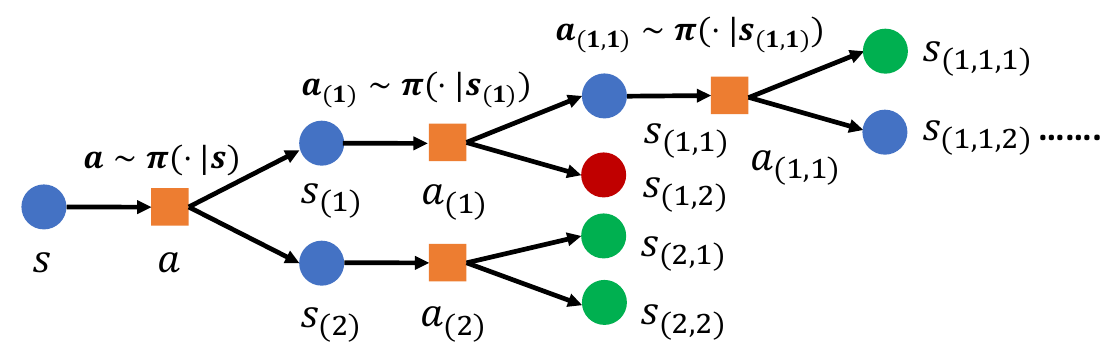}}
\caption{
An illustrative example of the tree-shaped MDP. 
Starting from the target molecule, chemists recursively choose reactions (denoted by orange rectangles) to break down the product molecules (denoted by blue circles) into reactants, until reaching building block molecules (denoted by green circles) or dead-end molecules (denoted by red circles).
In this example, the route is not synthesizable, as there is a red dead-end leaf node $S_{(1,2)}$. 
% Retrosynthesis aims to obtain synthesis routes that are 1) first synthesizable; 2) then as low-cost as possible. 
% The blue circle represents the target/intermediate molecule, the orange square represents the reaction chosen, the green circle represents the building block molecules and the red circle represents the dead-end molecule. 
% At each step, one reaction will be chosen and the product molecule will be broken into several reactants.
}
\label{fig:mdp}
\end{figure}

The task of retrosynthesis can be modeled as a 
% true-structured 
tree-shaped MDP $\mathcal{M} = (\mathcal{S}, \mathcal{A}, 2^\mathcal{S}, \mathcal{P}, c)$.
% for each molecule.
$\mathcal{S}$ denotes the state space whose element $s\in\mathcal{S}$ represents a molecule. Terminal states in $\mathcal{S}$ can be classified into (1) building blocks $\mathcal{S}_{\rm bb}$ which are commercially available molecules, and (2) dead-end molecules $\mathcal{S}_{\rm dead}$ to which no reactions are available. 
The element of the action space $a\in\mathcal{A}$ represents the chemical reaction that transforms a product molecule into reactant molecules. 
% The space of reactions are too large to deal with, in practice we use a reactions proposer $B: \mathcal{S}\rightarrow 2^\mathcal{A}$ that restricts the candidate reactions to a much smaller range.
The deterministic transition function $\mathcal{P}: \mathcal{S}\times\mathcal{A}\rightarrow 2^\mathcal{S}$ represents the single-step chemical reaction, whose inputs are the product molecule and the reaction to take, and the output is the set of reactants. 
Unlike a traditional MDP 
% where there is only a single successive state, 
where the trajectory
of each episode is a single path,
a trajectory of such an MDP forms a tree, since reactions usually have multiple reactants and thus cause branches (as illustrated in Fig.~\ref{fig:mdp}).
\textcolor{black}{The cost function $c: \mathcal{S}\times\mathcal{A}
\rightarrow \mathbb{R}$ consists of the cost of performing a certain reaction $c_{\rm rxn}(s,a)$ and the cost of
% penalizing for
reaching dead-end molecules $c_{\rm dead}(s,a)$ as follows:
% for which no reactions are possible:
}
\begin{equation}
c(s,a) = c_{\rm rxn}(s,a) + \underbrace{c_{\rm dead} \cdot \sum_{s' \in \mathcal{P}(s,a)}{ \mathbbm{1}(s' \in \mathcal{S}_{\rm dead})}}_{c_{\rm dead}(s,a)}.
\label{sec3:cost_function}
\end{equation}

\textcolor{black}{For example, \citep{simulated_experience} set $c_{\rm rxn}(s,a) = 1$ for all the reactions, $c_{\rm dead} =100$ for all dead-end molecules. Minimizing such cost function aims to generate routes that are synthesizable (i.e., no dead-end molecules in the route) and as short as possible. We note that our cost function is completely general, and can be trivially extended to for example account for building block or reagent prizes, or other route criteria such as convergence.
}

\textcolor{black}{
Given a policy function $\pi: \mathcal{S} \times \mathcal{A} \rightarrow [0, 1]$, the value function represents the expected total cost of the generated routes for molecule $s$:
}
\begin{equation}
    V_{\pi}(s) = \mathbb{E}_{\tau\sim\pi} \left[\sum_{(s', a')\in\tau} c(s', a') \right],
\label{sec3:total_cost}
\end{equation}
where $\tau$ is a tree-structured trajectory starting from state $s$.

% According to the definition, the value function also satisfies the Bellman equation

% \begin{equation}
%     V(s) = \mathbb{E}_{a\sim\pi(\cdot\mid s)}\left[c(s, a) + \sum_{s'\in\mathcal{T}(s, a)} V(s')\right] \, .
% \end{equation}

\paragraph{Dual Value Networks.}
% As we described above, the goal of retrosynthesis has two parts: 1) successful; 2) low-cost. 
As we can see, the desirable routes in retrosynthesis should be 1) synthesizable and 2) as low-cost as possible.
% \footnote{In this paper, low-cost means short.}. 
Instead of using one value network to capture both desiderata, we use the law of total expectation\footnote{$\text{E}[X] = \text{E}[\text{E}[X|Y]]$.} to decompose the function in Eqn.~\ref{sec3:total_cost} into two value functions of different kinds. Specifically,
% one stands for synthesizability, the other one stands for the reaction costs given synthesizability. 
let one random variable $X$ represent the total cost of the route: $\sum_{(s', a')\in\tau} c(s', a')$\footnote{Cost function $c(s,a)$ is defined in Eqn.~\ref{sec3:cost_function}.}, 
the other random variable $Y$ represent whether the route has no dead ends:  $\mathbbm{1}(\sum_{(s', a')\in\tau}{c_{\rm dead}(s', a') = 0)}$,
% ~\footnote{Intuitively, $Y$ means if the generated route $\mathcal{\tau}$ has dead-end molecules.}.
% so we have:
% so we can rewrite Eqn.~\ref{sec3:total_cost} as below:
then the value function in Eqn.~\ref{sec3:total_cost} can be rewritten as follows:
% ~\footnote{Omit the coefficient of $c_{\rm dead}$ here,  since $c_{\rm dead}$ is a relatively large penalty constant.}
\begin{equation}
\label{sec3:decomposition}
\begin{aligned}
    % &\mathbb{E}_{\tau\sim\pi} \left[\sum_{(s', a')\in\tau} c(s', a') \right] \\
    % &= \mathbb{E}[\mathbb{E}[\sum_{(s', a')\in\tau} c(s', a') | \mathbbm{1}(\sum_{(s', a')\in\tau}{c_{dead}(s', a') \ge c_{dead})} ]] \\
    % &= 
    % V(s) = \mathbb{E}[\mathbb{E}[\sum_{(s', a')\in\tau} c(s', a') | \mathbbm{1}(\sum_{(s', a')\in\tau}{c_{dead}(s', a') \ge c_{dead})} ]]
    &\mathbb{E}[X] = \mathbb{E}[\mathbb{E}[X|Y]] \\
    &= P_{\pi}(Y=1)\cdot\mathbb{E}[X|Y=1] +  P_{\pi}(Y=0)\cdot\mathbb{E}[X|Y=0] \\
    &\approx P_{\pi}(Y=1)\cdot\mathbb{E}\left[\sum_{(s', a')\in\tau} c_{\rm rxn}(s', a')|Y=1 \right] + \\ 
    &  P_{\pi}(Y=0)\cdot c_{\rm dead} 
    \quad \text{(omit the coefficient of $c_{\rm dead}$ here, } \\
    & \quad \text{since $c_{\rm dead}$ is a relatively large penalty constant).}
\end{aligned}
\end{equation}

One value network $V^{\rm syn}(s)$, called synthesizability value network, aims to approximate the probability term $P_{\pi}(Y=1)$. 
% Intuitively, it represents the probability of generating a successful route following current policy $\pi$. 
$V^{\rm syn}(s)$ represents the probability of generating a synthesizable route for molecule $s$
following policy $\pi$. 
% On the other hand, we can use
The other value network $V^{\rm cost}(s)$, called cost value network, aims to approximate the term $\mathbb{E}[\sum_{(s', a')\in\tau} c_{\rm rxn}(s', a')|Y=1]$. 
% Intuitively, it represents the expected reaction costs given that the route is successful. 
$V^{\rm cost}(s)$ represents the expected total reaction costs given the synthesizable route.
% generating a successful route 
% following current policy $\pi$. 
% Apart from the cost associated with the reactions, we can also utilize the estimation of the synthesizability of molecules as an auxiliary value function. Actually, these two values do not conflict but can work in harmony. When the confidence of synthesis is low, the cost estimation is not accurate while the synthesizability estimation is more reliable.
% The learning algorithm will leverage both value networks to generate simulated experiences and enhance the single-step model. 
% The two value networks will guide 
% Both networks guide the planning phase to generate both synthesizable and low-cost experiences.
Note that both value functions also satisfy the Bellman equation according to their mathematical definitions:
% The value function for the synthesizability is defined below
\begin{equation}
\begin{aligned}
V^{\rm syn}_{\pi}(s) & = \mathbb{E}_{a\sim\pi}\left[\prod_{s'\in\mathcal{P}(s, a)} V^{\rm syn}_\pi(s')\right], \\
V^{\rm cost}_{\pi}(s) & = \mathbb{E}_{a\sim\pi}\left[c_{\rm rxn}(s, a) + \sum_{s'\in\mathcal{P}(s, a)} V^{\rm cost}_{\pi}(s') |Y=1 \right].
\label{eqn:bellman}
\end{aligned}
\end{equation}
% We can see that the Bellman equation in a tree-shaped MDP is based on the product/sum over all children reactant nodes.
In a tree-shaped MDP, the Bellman equation is based on the product/sum over all children reactant nodes.

% due to the tree-shaped retrosynthesis MDP.
% due to the tree-shaped MDP.
% Since retrosynthesis MDP is tree-structured, the above bellman equation is based on the product/sum over all the reactant nodes. 
% The equation is useful for the backup step in section~\ref{sec:planning} and computing the training targets for dual value networks in section~\ref{sec:training}.

% \subsection{Planning for successful and low-cost routes}\label{sec:planning}
\subsection{Planning with Dual Value Networks}\label{sec:planning}

% To search synthetic routes on the retrosynthesis MDP, we proposed an MCTS-based planning algorithm that can fully leverage the strength of the dual value network. 
To learn optimal policies that lead to desirable routes on the retrosynthesis MDP, we propose an algorithm named Planning with Dual Value Networks (PDVN), which alternates between the planning phase and the updating phase. 

% During the planning phase. 
The planning phase aims to generate valuable simulated experiences on tree-shaped MDP with MCTS-based planning utilizing dual value networks. 
 % dual value networks.
First, we initialize an empty search tree with a target molecule as the root node.
In each iteration, a tree search process is executed from the current root node, utilizing the dual value networks and policy network. 
After the process completes, search probabilities based on the visit count of each reaction from the current root node are returned.
One reaction is chosen according 
to this search probabilities,
and the reactant nodes associated with the chosen reaction are pushed into a stack. 
% where the unsolved intermediate molecules are stored
% the search tree begins at a target molecule $s$, on which we do simulations that can explore the molecular space and back-propagate values to the simulation roots. After the simulations, we choose the most promising reaction according to the visitation frequencies and apply it to the simulation root.
% The resulting reactants are taken as the simulation roots for future iterations.
% We break down the molecule into a number of reactants through the reaction and the corresponding reactants are 
In the next iteration, a molecule is popped from the top of the stack and serves as a new root node. 
% to execute the tree search process.
The planning phase will conclude when either all the leaf nodes have been converted to building block molecules, or a dead-end leaf node is encountered.
% of the simulation, then we repeat the process of simulations, break down it into reactants and push them back to the stack.
% In this way, we are solving the target molecule by solving the intermediate molecules recursively until we completely break it down into the building blocks.
% We save all the MCTS search experiences excuated at each step in one search tree, which will be used to update the neural networks later.
% To be specific, 

In particular, each tree search process runs a predetermined number of simulations (e.g., 500 steps), and each simulation comprises three sequential steps: 1) Selection, 2) Expansion, and 3) Backup. We omit the step of rollout evaluation, and instead use networks to initialize the value of the newly expanded nodes, as this practice can effectively reduce the variance and computation efforts.

\paragraph{Selection.}
% To expand the search tree, we must first select a leaf node that can help solve the target molecule and has the potential to produce a low-cost synthesis route.
Starting from the current root node, we alternate between selecting a reaction node and a child molecule node, until we encounter a leaf molecule node.
% Since the search tree interleaves molecule layers with reaction layers, we need to devise selection rules for both cases.
To select reaction nodes,
we propose a modified version of the PUCT rule~\cite{rosin2011multi} that considers dual value networks.
The new rule includes an estimate of synthesizability $R(s, a)$, an estimate of cost $Q(s, a)$, the policy $\pi(a|s)$, and the visit count $N(s, a)$;
% For the molecule layers, 
% we select reactions by following the PUCT rule. For the candidate reaction node $a$ of the molecule node $s$, we keep track of the dual value estimates, i.e., the cost estimates $Q(s, a)$ and the prior estimates $R(s, a)$, as well as the visit counts $N(s, a)$ and the model prior $P(s, a)$.
% To combine the dual value estimates, we define a new value proxy that is consistent with the Eqn.~\ref{sec3:decomposition},
% Specifically, the following 
its detailed equation 
% selection rule 
is derived from Eqn.~\ref{sec3:decomposition}:
% \begin{equation}
% \end{equation}
% Given the exploration constant $C$, the reaction is selected according to the following rule:
\begin{equation}
\begin{aligned}
&a = \argmax_{a'} -U(s, a') + C \, \pi(a'|s) \, \frac{\sqrt{\sum_{b}N(s, b)}}{1 + N(s, a')}, \\
&U(s, a') = R(s, a') \cdot Q(s, a') + (1 - R(s, a')) \cdot c_{\rm dead},
\label{eq:puct}
\end{aligned}
\end{equation}
where $C$ is the exploration coefficient. 
% We can see that when $R(s, a')$ is close to $1$, which means reaction $a'$ is very likely to derive a synthesizable route, $U(s, a')$ will pay more attention to the cost estimation $Q(s, a')$. 
% Otherwise, $U(s, a')$ will pay more attention to the penalty constant $c_{\rm dead}$.
To select a child molecule node\footnote{Since each reaction node may have multiple child molecule nodes in a tree-shaped MDP},
we prioritize molecules that have not been expanded;
% we give priority to those child molecules that have not been expanded. 
if none are available, we choose ones that have not been solved. 
% If all molecules have been either expanded or solved, we randomly select one.
If molecule nodes are either all expanded or all solved, we randomly select one of them.
% If all child molecules have been expanded, we select those child molecules that have not been solved. 
% If all child molecules have been solved, we randomly select one.
% Otherwise, all the molecules are solved before and then we randomly select one in order to further optimize the synthesis routes of the solved molecules.

\paragraph{Expansion.}
When a leaf molecule node is encountered, we expand the search tree by adding reaction nodes and their corresponding reactant nodes. 
Specifically, we select the top 50 predictions of the policy network to append reaction nodes\footnote{Since 50 is widely used in previous work (e.g., 3N-MCTS, Retro*, RetroGraph, Retro*+). 
% According to \cite{}
% The first related paper to use 50 is [ref1], since they find 1) The correct reactions are generally ranked highly (e.g., top-50 accuracy is about 72.5%). 2) Beyond the top 50 predicted results, accuracy increases only marginally. So, they use 50 to effectively reduce the branching factor drastically, which is 46,175 when rules are applied exhaustively.}
}, and then use RDKit\footnote{https://www.rdkit.org/, open-source cheminformatics software.} to obtain the corresponding reactant nodes. 
% , both the leaf node and the reaction are passed to  to obtain the corresponding reactant nodes. 
For new molecule nodes, the visit count is initially set to zero; the dual value networks assign values for the initial $V^{\rm syn}$ and $V^{\rm cost}$. 

% are set by the dual value networks.     
% These reactant nodes are the child nodes of the reaction node.   
% we expand it so that the search tree grows. A reaction proposer is used to propose candidate reactions $a\in B(s)$ for the leaf node. The leaf node $s$ and the candidate reaction $a$ are then passed to the software to generate the reactants $s'\in\mathcal{T}(s, a)$.
% We attach the reaction nodes $r\in B(s)$ to the leaf node $s$ and the molecule nodes $s'\in\mathcal{T}(s, a)$ to the reaction node $r$.

\paragraph{Backup.}
At the end of each simulation, the nodes visited, both the reaction and molecule nodes, form a path $T = (s_0, a_0, \dots, s_l, a_l, \dots, s_L)$, where $s_0$ is the root of the current simulation and $s_L$ is the leaf node before expansion.
% starting from  current root node $s_1$. 
% to the expanded node $s_L$.
% The statistics associated with each node in the path are updated in the backward direction.
During the backup step, we first calculate the current values of each molecule node $s_l$ ($ 0 \le l \le L-1$ ) on path $T$ by: 
% (e.g., $V_T^{syn}(s_l)$, $V_T^{cost}(s_l)$) 
% according to Eqn.~\ref{eqn:bellman}. 
% \begin{equation}
% \begin{aligned}
% V_{T}^{syn}(s_l) = xxx. \\
% V_{T}^{cost}(s_l)
% \end{aligned}
% \end{equation}
\begin{equation}
\begin{aligned}
V_{T}^{\rm syn}(s_l) & = V_T^{\rm syn}(s_{l+1}) \cdot \prod_{s'\in\mathcal{P}(s_l, a_l)\setminus\{s_{l+1}\}} V^{\rm syn}(s'), \\
V_{T}^{\rm cost}(s_l) & = c_{\rm rxn}(s_l, a_l) + V_T^{\rm cost}(s_{l+1})  + \\
&\sum_{s'\in\mathcal{P}(s_l, a_l)\setminus\{s_{l+1}\}} V^{\rm cost}(s'). \\
\label{eqn:backup}
\end{aligned}
\end{equation}
The above update rule is derived from the Bellman equation for a tree-shaped MDP (i.e., Eqn.~\ref{eqn:bellman}).  
Then, we update the average value
$V^{\rm syn}(s_l)$, $V^{\rm cost}(s_l)$ by $V^{\rm syn}(s_l) = (V^{\rm syn}(s_l) \cdot N(s_l) + V_T^{\rm syn}(s_l)) / (N(s_l) + 1)$, $V^{\rm cost}(s_l) = (V^{\rm cost}(s_l) \cdot N(s_l) + V^{\rm cost}_{T}(s_l)) / (N(s_l) + 1)$,  and the visit count by $N(s_l) = N(s_l)+1$.
% The dual values of the expanded node $s_L$ are initialized as $V_T(s_L)=v_\theta(s_L)$ and $R_T(s_L)=r_\theta(s_L)$. 
% Then, the dual values are calculated recursively along the path in the following manner:
Finally, we update the values of reaction nodes as follows:
\begin{equation}
\begin{aligned}
R(s_l, a_l) & = \prod_{s'\in\mathcal{P}(s_l, a_l)} V^{\rm syn}(s'), \\
Q(s_l, a_l) & = c_{\rm rxn}(s_l, a_l) + \sum_{s'\in\mathcal{P}(s_l, a_l)} V^{\rm cost}(s'). \\
\end{aligned}
\end{equation}
% where $V_T(s_l) = V^{cost}_{T}(s_l, a_l)$, $R_T(s_l) = V^{syn}_T(s_l, a_l)$ for $l < L$. 
Note that $V_T^{\rm syn}(s_l)$ and $V_T^{\rm cost}(s_l)$ denote the values calculated from current path $T$, while $V^{\rm syn}(s_l)$ and $V^{\rm cost}(s_l)$ denote the average values over all visited paths.
% $V(s_l)$ and $R(s_l)$ are average dual values over these path values:
% Finally, we increment the visit count $N(s_l)$ by one for all visited molecule nodes.

\subsection{Training on Generated Experiences}\label{sec:training}
After completing the Planning phase, we have a search tree along with the statistics gathered during planning. 
This tree contains valuable information for updating the networks, regardless of whether it solves the target molecule, helping to benefit future planning.
% Regardless of whether the search tree solves the target molecule, this search tree contains valuable information for updating the networks, so as to benefit future planning. 
% The search tree contains useful data that can benefit future planning, no matter if the search tree solves the target molecule or not. 
% However, the three networks take different roles during planning and the training targets are quite different for each of them. 
During PDVN training, three neural networks (i.e., policy network, synthesizability value network, and cost value network) play different roles. 
% during PDVN training. 
In this subsection, we carefully design the process of extracting data and the objective function for training each network.

% \paragraph{Single-Step Model.}
\paragraph{Policy Network.}
% The single-step model is trained to imitate the successful routes in the search tree, so that the gap between single-step and multi-step retrosynthesis can be narrowed.
% To narrow the gap between single-step and multi-step retrosynthesis, the single-step model is trained to imitate the multi-step planner.
The policy network $\pi(a|s)$ aims at predicting the reactions that lead to desirable routes.
% Instead of maximizing the similarity of the policy network probabilities $\pi$ to the visitation frequency $N(s, \cdot)$ outputted by MCTS simulations, as in AlphaZero~\citep{silver2017mastering, silver2018general}, 
Instead of imitating the visitation frequency $N(s, a) / \sum_b N(s, b)$ from MCTS simulations as in AlphaZero~\citep{silver2017mastering, silver2018general}, 
we extract pairs of (molecule, reaction) from successful routes with minimal cost in the search tree.
% we only predict the successful reactions that solve the molecules given by the planner for two reasons: (1) successful reactions incorporate the signal of synthesizability into the dataset (2) visitation frequency is too noisy when the number of simulations is not large enough.
Specifically, for each molecule node in the search tree, we first determine if there are any reactions leading to a successful route.
If more than one reaction meet the required condition, we will select the one with the lowest cost. %/length.
We use cross-entropy (CE) loss to update the policy network.

% given a search tree that solves the target molecule, we collect the molecule-reaction pair $(s, a)$ if the molecule $s$ is solved by the reaction $a$. If there are multiple reactions that could solve the molecule, we choose the one with the minimal cost. Then the single-step model is trained on the collected molecule-reaction pairs by the cross-entropy loss.
% The synthesizability in the retrosynthesis task provides useful guidance to extract training data for cross-entropy (CE) loss.

\paragraph{Synthesizability Value Network.}
The synthesizability value network $V^{\rm syn}(s)$ aims to predict the probability of solving molecule $s$. 
% Unlike the single-step model, 
We train $V^{\rm syn}(s)$ by using all the molecules in the search tree. 
First, we run a recursive algorithm to check if each molecule node in the search tree is solved or not. For solved molecule nodes, we set the training target as $1$. 
For unsolved molecules $s$, we use $0.8 \times V^{\rm syn}(s)$ as the training target, where $0.8$ is a slight penalty since the molecule is not solved in the search tree.
% \footnote{$\alpha = 0.8$ works well empircially.}.  
For dead-end molecules, we set the training target as $0$. 
We use binary cross-entropy (BCE) loss to update the synthesizability value network.
% For the solved molecule node, we set the target to $1$. 

\paragraph{Cost Value Network.}
The cost value function $V^{\rm cost}(s)$ aims to estimate the minimal cost or length of synthesizing the molecule. 
$V^{\rm cost}(s)$ is trained only on solved molecules in the search tree.
Specifically, we first use a recursive algorithm to obtain the lengths of the shortest successful routes on those solved molecules, which we use as the training target for $V^{\rm cost}(s)$. 
We minimize the mean squared error (MSE) loss to update the cost value network.
% and the training target is the minimal cost of solving the molecule. 
% The loss function for the cost value network is chosen to be the mean squared error.

\subsection{Two-Branch Policy Network Structure}\label{sec:realistic} 

% \begin{figure}[h]
% \centerline{\includegraphics[width=0.5\textwidth]{realistic policy.png}}
% \caption{Illustration of two-branch policy network structure.} 
% \label{fig:realistic}
% \end{figure}

\begin{figure}[t]
\centerline{\includegraphics[width=0.5\textwidth]{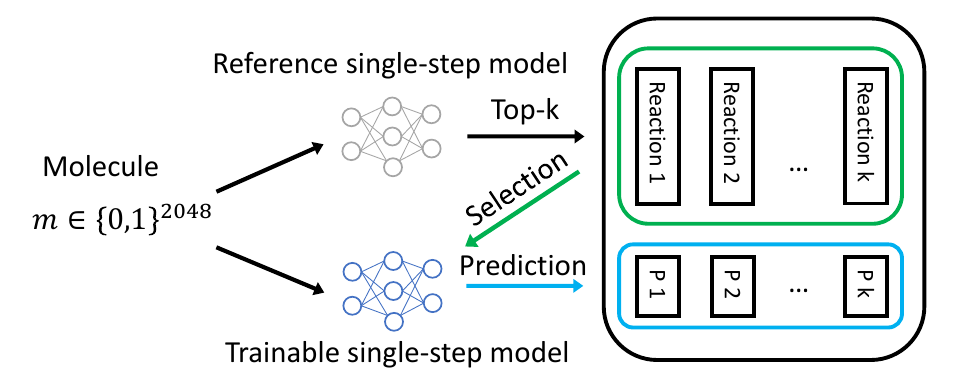}}
\caption{An illustration of the two-branch policy network. The reference single-step model provides a realistic subset of reactions for the input molecule, denoted by Reaction $1$ \dots, Reaction $k$ . 
The learnable single-step network optimizes a probability distribution over the selected reactions, i.e., $P_{i}$ is the probability of $\text{Reaction } i$.}
\label{fig:two-branch}
\end{figure}

% \textbf{\#figure illustrating realistic policy structure}.

The above subsections focus on optimizing policies to generate desirable routes. 
In this subsection, we focus more on how to retain single-step accuracy.

A natural way to design a policy network is to directly use a single-step model. However, as the training proceeds, 
% such policy may choose unrealistic reactions that can pass RDKit but are not likely to happen in practice. 
such policy may choose unrealistic reactions that are not likely to happen in practice. 
Chemists often question the feasibility of the routes generated by AI-based retrosynthesis software~\cite{genheden2022paroutes}.
% To address this issue, 
To this end, 
we design a two-branch policy network structure, as illustrated in Fig.~\ref{fig:two-branch}.  

Specifically, the proposed policy network consists of two separate branches. 
The first branch, called the reference single-step model, is inherited from a single-step model trained by offline SL, and frozen during RL training. 
Following \citep{Segler2018PlanningCS, chen20retrostar}, we use a template-based MLP model as the single-step model, which is a multi-class classification network\footnote{Template-based approaches use reaction templates to predict reactants from a product. First, the template-based model predicts the reaction templates, and then the template is applied to a product to find a match via subgraph isomorphism. If a proper isomorphism is found, the product is transformed according to the template.}.
Since the model is pre-trained using the real-world reaction dataset, 
this branch provides a realistic subset of reactions (e.g., the top-$50$ candidates from a total of $\sim 380K$ classes).
% which casts single-step retrosynthesis as a multi-class classification problem~\footnote{Given a molecule as product, the goal is to predict possible reaction templates, reactants are obtained by applying the predicted templates to product molecule.}.
% ~\citep{chen20retrostar}.
% To be specific, we feed the molecule into two networks with the same structure mapping from the molecule fingerprint to the distributions over templates. The two branches have different but complimentary functionality:
% One branch offers the set of reaction candidates by selecting the reactions within the top-50 probabilities. The network is usually pretrained on a single-step retrosynthesis dataset.
% The second branch is initialized by the SL model but has  learnable parameters to optimize the probability over the provided reaction set for generating synthesizable and low-cost routes. 
% Two branches complement each other to generate both realistic and desirable synthesis routes.
The second branch is initialized by the SL model, but has learnable parameters that can be optimized to generate synthesizable and cost-effective routes from the realistic subset of reactions. 
The training objective refers to the Policy Network part in section~\ref{sec:training}.
These two branches work together to generate both realistic and desirable synthesis routes.

% On the other hand, the second branch labels the probability of each reaction, and the parameters are usually initialized to be the same as the first branch.
% During training, we only update the parameters of the second branch so that the reaction candidates for a molecule is fixed but the probability is changed to suit the purpose of multi-step retrosynthesis.

\section{Experiments}
% 1.4. Di's version
% In this section, we design experiments to answer the following questions: \textbf{Q1:} How does the training method improve the accuracy and the route quality on test dataset? \textbf{Q2:} How does the trained one-step model perform on other datasets? \textbf{Q3:} Is the dual value network construction necessary for the performance? \textbf{Q4:} Can we directly train the model on harder molecules? \textbf{Q5:} Is the synthesis routes given by the trained one-step model practical for professions?

\begin{table*}[t]
    \caption{Performance summary on the USPTO 190 test dataset. The evaluation metrics include the success rates under different numbers of model calls ($N$), the average number of model calls used, the average number of reaction nodes (T) and molecule nodes (M) visited, under the computation budget of $500$ model calls.}
    \label{tab:main_results}
    \centering
    \resizebox{\textwidth}{!} {%
    \begin{tabular}{l cccccc ccc}
        \toprule
         & \multicolumn{6}{c}{Success rate $[\%] \uparrow$} & \\
        \cmidrule(lr){2-7} 
        Algorithm & 50  & 100 & 200 & 300 & 400 & 500 & \# model calls $\downarrow$ & \# T nodes $\downarrow$ & \# M nodes $\downarrow$ \\
        \midrule
        Greedy DFS & - & 38.42 & 40.53 & 44.21 & 45.26 & 46.84 & 300.56 & - & - \\
        DFPN-E & - & 50.53 & 58.42 & 64.21 & 68.42 & 75.26 & 208.12 & 3123.33 & 4635.08 \\
        MCTS-rollout & - & 43.68 & 47.37 & 54.74 & 58.95 & 62.63 & 254.32 & - & - \\
        \midrule
        Retro*-0 & 27.37 & 38.42 & 58.42 & 67.37 & 75.26 & 79.47 & 209.86 & 3905.62 & 5565.37 \\
        Retro* & 40.00 & 55.79 & 70.53 & 76.84 & 82.11 & 85.79 & 158.81 & 2632.84 & 3685.31 \\
        Retro*+-0 & 56.84 & 67.37 & 83.16 & 92.11 & 94.74 & 96.32 & 97.95 & 1444.52 & 2139.3 \\
        Retro*+ & 63.16 & 74.21 & 83.16 & 86.84 & 90.00 & 90.53 & 98.91 & 1157.74 & 1708.17 \\
        % \textbf{Ours+Retro* (V2)} & 82.11 & 88.42 & 93.68 & 95.79 & 97.89 & 98.42 & 42.26 & 921.48 & 1174.71 \\\
        % EG-MCTS & - & 85.79 & 92.63 & 94.21 & 95.79 & 96.84 & 55.84 & 869.59 & 1193.79 \\ (Cannot run open source code. Not published.)
        \textbf{PDVN+Retro*-0} & \textbf{86.32} & \textbf{93.68} & \textbf{97.37} & \textbf{97.89} & \textbf{98.95} & \textbf{98.95} & \textbf{30.94} & \textbf{773.56} &  \textbf{995.22} \\
        \midrule
        RetroGraph & 69.47 & 88.42 & 97.89 & 98.95 & 99.47 & 99.47 & 45.13 & 674.23 & 500.44 \\
        % \midrule
        % \textbf{Ours+Retro*} & 63.68 & 79.47 & 88.95 & 93.16 & 95.79 & 96.32 & 75.69 & 1477.91 & 1988.64 \\
        % \textbf{Ours+RetroGraph} & 80.00         & \textbf{93.16}         & 97.89         & 5.69              \\
        % \textbf{Ours+RetroGraph} & 85.79 & 92.11 & 95.79 & 97.89 & 98.42 & 99.47 & 39.60 & 807.38 & 609.03 \\
        % \textbf{Ours+Retro* (V2)} & 88.95 & 94.21 & 97.37 & 97.37 & 98.42 & 98.95 & 28.68 & 666.92 & 856.80 \\
        % \textbf{Ours+online MCTS (V2)} & 82.11 & 87.89 & 91.58 & 95.26 & 95.79 & 96.84 & 51.10 & 963.96 & 1291.62 \\
        % \textbf{Ours+RetroGraph (V2)} & 87.89 & 93.68 & 95.79 & 97.37 & 97.89 & 97.89 & 35.07 & 835.82 & 705.83 \\
        \textbf{PDVN+RetroGraph} & \textbf{93.16} & \textbf{96.84} & \textbf{97.89} & \textbf{99.47} & \textbf{99.47} & \textbf{99.47} & \textbf{20.24} & \textbf{486.87} & \textbf{417.54} \\
        
        % \scalebox{0.8}{\> ($\text{temperature} = 1.6$)} & \\
        \bottomrule
    \end{tabular}%
    }
    % \caption{Experimental results on USPTO 190 test dataset. }
\end{table*}

\textcolor{black}{In this section, we aim to answer the following questions: \textbf{Q1:} On the widely-used USPTO dataset, can our PDVN algorithm significantly improve the performance of existing multi-step planners?
\textbf{Q2}: On more test target molecules from ChEMBL and GDB-17, can PDVN algorithm still bring performance gains?
\textbf{Q3}: Are the proposed dual value networks necessary in our algorithm? 
\textbf{Q4}: 
% How does the single-step model trained by PDVN achieve better results, compared with the SL model?
Does PDVN algorithm helps find chemically sound routes?
% Does the multi-step planner equipped with the  single-step model trained by our method generate chemically reasonable routes?
% \textbf{Q5}: Does our method generate chemically reasonable routes? case study. 
}
\subsection{Experimental Setup}
% Before answering the questions, we first 
Our algorithm requires specifying (1) a set of building block molecules $\mathcal{S}_{bb}$,
(2) a training target molecule dataset $\mathcal{D}_{\rm train}$,
(3) a test target molecule dataset $\mathcal{D}_{\rm test}$,
(4) a retrosynthetic planning algorithm, 
(5) a single-step model.

\paragraph{Dataset.}
% \textbf{1. building blocks. 2. training target molecules. 3. test target molecules. }
(1) For the building block molecules $\mathcal{S}_{bb}$, we follow common practice~\cite{chen20retrostar}, and use the commercially available molecules (about $23.1M$) from \textit{eMolecules}\footnote{https://downloads.emolecules.com/free/2022-11-01/}.
% In practice, chemists can choose any set of molecules according to their own application scenarios. 
(2) For the training target molecule dataset $\mathcal{D}_{\rm train}$, we follow the procedure from \cite{chen20retrostar, kim2021self} and construct synthesis routes based on the publicly available reaction dataset extracted from the United States Patent Office (USPTO)~\cite{lowe2012extraction}. 
% and building blocks from \textit{eMolecules}.
% eMolecules consists of 23.1M~\footnote{But we find the size of building blocks is about 23.1M in their codebase.} commercially available molecules.
Specifically, they take each molecule that has appeared in USPTO reaction data and analyze if it can be synthesized by existing reactions within USPTO training data. After processing, $299202$ training routes are obtained. 
Following \citep{kim2021self}, we use the root molecules of these training routes as training target molecules to generate simulated experiences in the Planning phase.
% For each synthesizable molecule, we choose
% the shortest-possible synthesis routes with ending points
% being available building blocks in eMolecules.
(3) For the test target molecules $\mathcal{D}_{\rm test}$, we use the $190$ challenging target molecules that were widely used in previous work~\cite{chen20retrostar, kim2021self, han2022gnn, xie2022retrograph, tripp2022reevaluating}.
Besides, we introduce two novel test datasets, i.e., ChEMBL-1000 and GDB17-1000, to assess the generalizability of the trained model.

\paragraph{Multi-Step Planner.} We use two popular multi-step planners: 1) Retro*~\cite{chen20retrostar}, an efficient retrosynthetic planning algorithm built upon the AND-OR search tree and best-first search;
% 2) RetroGraph~\cite{xie2022retrograph}, which reduces  duplication in the AND-OR search tree and search in the AND-OR graph.
RetroGraph~\cite{xie2022retrograph}, which proposes to use AND-OR graph to handle the duplicated nodes and searches within the retrosynthetic paths.
% and
% RETRO*-0 (Chen et al., 2020b) as the search algorithm
Note that Retro*-0 denotes a variant of Retro* not relying on the pretrained value function as the heuristic~\citep{chen20retrostar}. 
% which estimates the cost to synthesize
% a given molecule, proposed by Chen et al. (2020b).
These two planners are based on the A* algorithm and can be combined with any single-step model. 
We plug the single-step model trained by PDVN into these two planners to see if there are any improvements on the test target molecule dataset. 

\paragraph{Single-Step Model.}
% As one branch of the proposed two-branch policy network (in section~\ref{sec:realistic}), the reference single-step model trained by SL is critical for the policy network to predict valid reactions.  
Following~\cite{Segler2018PlanningCS, chen20retrostar, kim2021self}, we use a template-based single-step model, which is a 2-layer MLP with ELU activation. The output layer has $\sim380K$ units and each corresponds to a distinct template.
Although simple, this model is adopted in the most widely used retrosynthesis implementations (e.g., Retro*, ASKCOS, and AIZynthfinder). 
To train this model, we follow~\cite{chen20retrostar} and use the offline training dataset comprising $\sim 1.3M$ reactions extracted from USPTO published up to September 2016.
% There are in total $\sim380K$ distinct templates.
% % so the output size of the MLP is $\sim380K$.
% The offline training dataset 

\paragraph{Dual value networks.}
For the synthesizability value network, we use a 2-layer MLP where the output layer is a sigmoid layer.
The cost value network uses the same network architecture but the output layer is a softplus layer as the cost of non-building blocks is positive.
% The input to both networks is the molecular Morgan fingerprint feature,
% The input feature to both of the networks is the Morgan fingerprint of a molecule, 
% which is a $2048$-bit vector and of radius $2$.
The input to both networks is the molecular Morgan fingerprint of radius 2, which is a 2048-bit vector.

\paragraph{Training.}
For the two-branch policy network, the reference model is pre-trained offline by SL and then frozen during PDVN training. 
The learnable branch has the same network architecture as the reference model and is initialized by the reference model. 
% but the network parameters are learnable during PDVN training.
Instead of training the reference model from scratch, we load the model checkpoint provided by \cite{chen20retrostar}.
The parameters of the dual value networks are randomly initialized.
During the Planning phase, the batch size of sampled target molecules is $1024$.
We set $c_{rxn}(s, a) = 0.1$ and $c_{dead} = 5$. 
During the Updating phase, the Adam optimizer~\cite{kingma2014adam} with a mini-batch of size $128$ and a learning rate of $0.001$ is used  for all models.
% The training epoch number is $8$.
We iterate the training target molecule dataset $\mathcal{D}_{\rm train}$ three times.

\subsection{USPTO Results}

\begin{table}[t]
    \caption{The average length of the routes on the USPTO 190 test dataset. The results are averaged over the $138$ molecules that can be solved by all methods.}
    \label{tab:quality_metrics}
    \centering
    \begin{tabular}{l c c}
        \toprule
        %  & \multicolumn{6}{c}{Success rate $[\%] \uparrow$} & \\
        % \cmidrule(lr){2-7} 
        Algorithm & Avg length $\downarrow$ \\
        % & Avg  cost $\downarrow$ \\
        \midrule
        Retro*-0 & 5.83 \\ 
        % & 6.67 \\
        Retro* & 5.76 \\
        % & 6.92 \\
        Retro*+-0 & 6.16 \\
        % & 7.67 \\
        Retro*+ & 5.77 \\
        % & 7.78 \\
        % \textbf{Ours+Retro* (V2)} & 5.04 \\
        \textbf{PDVN+Retro*} & \textbf{4.83} \\
        % & 23.0 \\
        \midrule
        RetroGraph & 5.63 \\
        % & - \\
        % \textbf{Ours+Retro*} & 5.55 & 8.59 \\
        % \textbf{Ours+RetroGraph} & 80.00         & \textbf{93.16}         & 97.89         & 5.69              \\
        % \textbf{Ours+RetroGraph} & 4.94 & - \\
        % \scalebox{0.8}{\> ($\text{temperature} = 1.6$)} & \\
        % \textbf{Ours+online MCTS (V2)} & 4.88 & 21.98 \\
        \textbf{PDVN+RetroGraph} & \textbf{4.78}\\ 
        % - & - \\
        \bottomrule
    \end{tabular}
\end{table}

% \textbf{1. use which planners? 2. use PDVN trained model v.s. SL model. 3. Evaluation metric. 4. Results. Improve both planners. show number, SOTA, 10x less model calls. Discuss Retro*+. length as Route quality.}
To answer \textbf{Q1}, we investigate the effectiveness of PDVN in terms of planning efficiency and route quality. We train the single-step model by PDVN and load the trained model to two state-of-the-art retrosynthesis planners, i.e., Retro* and RetroGraph. The results against other baselines are summarized in Table~\ref{tab:main_results} and Table~\ref{tab:quality_metrics}.

We can observe that the success rates of both Retro* and RetroGraph increase significantly when combined with the PDVN trained model, and the average number of model calls is reduced by more than half. With the model calls limit $N=500$, the success rate of Retro* is improved to $98.95\%$ while Retro*+, which imitates the successful routes given by Retro*, achieves a success rate of $96.32\%$. Notably, PDVN improves the success rate for fewer model calls limit, especially $N=50$. 
% Previous to our method, the best result 
Previous best result with $N=50$ is $69.47\%$ by RetroGraph, PDVN significantly improves it to $93.16\%$.

On the other hand, we use the average route length to represent route quality.
% We mind the readers that previous work may use a large number, e.g., 32, as the length of unsolved molecules, which is misleading because the resulting average length is highly correlated to the success rate.
To make a fair comparison, we only consider the molecules that can be solved by all the planners. As shown in  Table~\ref{tab:quality_metrics}, PDVN outperforms the baselines by a large margin: the trained model reduces the average route length on the $138$ selected molecules from $5.83$ to $4.83$ for Retro*-0 and from $5.63$ to $4.78$ for RetroGraph.

% \paragraph{Results on ChEMBL-1000 and GDB17-1000}. 
\subsection{Results on ChEMBL and GDB-17 Datasets}

% \textbf{1. how we get those two datasets in details, mention Austin's paper.
% 2. observe the performance gains. GDB17 is indeed harder but improvement is also observed.}
% The USPTO test dataset contains a small number of molecules and nearly all test molecules are solved by RetroGraph when using $500$ model calls. The USPTO test dataset may not fulfill the purpose of a benchmark that can help analyze the all-around performance of SOTA methods.
% Due to this concern, we construct two datasets, i.e., ChEMBL-1000 and GDB17-1000, each of which contains 1000 molecules, posing new challenges to the retrosynthesis planners.

% Since the USPTO test dataset is a highly curated dataset~\citep{chen20retrostar}, e.g., the molecules are selected as those can be solved by DFS algorithm and the reactions are known to be within the top-50 probabilities of the pretrained one-step model.
% In other words, the complexity of the USPTO test dataset is quite limited.

% The USPTO test dataset contains a small number of molecules and nearly all the molecules are solved by RetroGraph when using $500$ model calls. The USPTO test dataset may not fulfill the purpose of a benchmark that can help analyze the performance of SOTA methods.

To answer \textbf{Q2}, we follow \cite{tripp2022reevaluating} and construct two more test datasets from the ChEMBL dataset and GDB17 dataset, i.e., ChEMBL-1000 and GDB17-1000. 
They are 1000 molecules randomly chosen from a subset of the ChEMBL dataset and GDB17 dataset.
% from  with the reaction model and purchasable molecules 
% from~\citep{chen20retrostar},
To create a subset of molecules equally or more challenging than the USPTO 190 test dataset, 
we preprocess the CHEMBL dataset and GDB17 dataset by using the script from~\cite{brown2019guacamol}, 
% the subset of molecules from ChEMBL-1000 was chosen by applying the pre-processing, 
keeping only molecules whose 
molecular weight, Bertz coefficient, logP, and TPSA were larger than the mean of the respective values in the USPTO 190 test dataset, and removing all known building block molecules.
%  the subset of
% molecules was chosen by applying the pre-processing script from [Brown et al., 2019],
% keeping only molecules whose molecular weight, Bertz coefficient, logP and TPSA were
% larger than the mean of the respective values in the Retro* hard set, and removing all
% purchasable molecules.

% \begin{enumerate}
%     \item For ChEMBL-1000, we randomly chose 1000 “hard” molecules from a subset of the ChEMBL dataset, with the reaction model and purchasable molecules from~\citep{chen20retrostar}. To create a set of molecules equally or more challenging than the Retro* set, the subset of molecules was chosen by applying the pre-processing, keeping only molecules whose molecular weight, Bertz coefficient, logP and TPSA were larger than the mean of the respective values in the Retro* hard set, and removing all purchasable molecules.
%     \item GDB17-1000 is composed of 1000 randomly chosen molecules from gdb17, using the reaction model and set of purchasable molecules from~\citep{chen20retrostar}.
% \end{enumerate}

% Since these two datasets are much harder, we only compare the success rate with the computation budget of model calls $N=500$. 
The results are reported in Table~\ref{tab:own_datasets_results}, and we can see that PDVN still brings performance gains to the baselines. For ChEMBL-1000, PDVN+Retro*-0 and PDVN+RetroGraph can solve $24$ and $8$ more molecules than Retro*+-0 and RetroGraph, respectively. The GDB17-1000 dataset is much harder but the improvement is more obvious. Retro*+-0 and RetroGraph can solve $154$ and $215$ molecules, and our method enables both planners to additionally solve more than half of what they can originally do, solving as many as $269$ and $371$ molecules, respectively.

\begin{table}[t]
\caption{Number of solved target molecules on ChEMBL-100 dataset and GDB17-1000 datasets.}
\label{tab:own_datasets_results}
\centering
\resizebox{\columnwidth}{!}{%
\begin{tabular}{l c c}
    \toprule
    %  & \multicolumn{6}{c}{Success rate $[\%] \uparrow$} & \\
    % \cmidrule(lr){2-7} 
    Algorithm & ChEMBL-1000 & GDB17-1000 \\
    \midrule
    Retro*-0 & 751 & 75 \\                          
    Retro* & 762 & 95 \\
    Retro*+-0 & 811 & 150 \\
    Retro*+ & 818 & 154 \\
    % \textbf{Ours+Retro* (V2)} & 853 & 325 \\
    \textbf{PDVN+Retro*-0} & \textbf{835} & \textbf{269}  \\
    \midrule
    RetroGraph & 852 & 215 \\
    % \textbf{Ours+Retro*} & 791 & 123 \\
    % \textbf{Ours+RetroGraph} & - & - \\
    \textbf{PDVN+RetroGraph} & \textbf{860} & \textbf{371} \\
    \bottomrule
\end{tabular}%
}
% \caption{Experimental results on USPTO 190 test dataset. }
\end{table}

\subsection{Ablation Study on Dual Value Networks}

\begin{table}[t]
\caption{Ablation study on dual value networks. 
% Use the USPTO 190 test dataset, and the route length is averaged over $138$ molecules that are solved by all the methods.
}
\label{tab:ablation_study}
\centering
\resizebox{\columnwidth}{!} {%
\begin{tabular}{l c c}
    \toprule
    %  & \multicolumn{6}{c}{Success rate $[\%] \uparrow$} & \\
    % \cmidrule(lr){2-7} 
    Algorithm & Success rate & Avg length \\
    \midrule
    % Retro*-0 & 751 & 75 \\                          
    % Retro* & 762 & 95 \\
    % Retro*+-0 & 811 & 150 \\
    % Retro*+ & 818 & 154 \\
    % \textbf{Ours+Retro* (V2)} & 853 & 325 \\
    \textbf{PDVN+Retro*-0} & \textbf{98.95} & \textbf{4.83}  \\
    SingleValue+Retro*-0 & 95.26 & 5.05  \\
    % Synthesizability+Retro*-0 & 95.79 & 5.16  
    PDVN w/o Cost+Retro*-0 & 95.79 & 5.16  \\
    \midrule
    % RetroGraph & 852 & 215 \\
    % \textbf{Ours+Retro*} & 791 & 123 \\
    % \textbf{Ours+RetroGraph} & - & - \\
    \textbf{PDVN+RetroGraph} & \textbf{99.47} & \textbf{4.78} \\
    SingleValue+RetroGraph & 96.32 & 4.93  \\
    PDVN w/o Cost+RetroGraph & 96.32 & 5.00  \\
    \bottomrule
\end{tabular}%
}
% \caption{Experimental results on USPTO 190 test dataset. }
\end{table}

In order to verify the necessity of the proposed dual value networks, we design two variants of PDVN using only one value network for comparison. 
For the first variant, which we call SingleValue, we use a single value network $V^{\rm single}(s)$ to estimate the overall objective of retrosynthesis in Eqn.~\ref{sec3:total_cost}.
% This value is backed up like the cost value $V^{\rm cost}(s)$ in Eqn.~\ref{eqn:backup}. 
The backup step of $V^{\rm single}(s)$ is similar to that of  the cost value $V^{\rm cost}(s)$ described in Eqn.~\ref{eqn:backup}.
% but is trained on all the non-leaf molecules in the search tree. 
The PUCT rule is also modified to use the single value $V^{\rm single}(s)$.
% calculated from the cost defined in Eqn.~\ref{sec3:total_cost} 
% and the initialized value given by $V^{\rm single}(s)$ as the value.
For the second variant, which is called PDVN w/o Cost, we remove the cost value network in PDVN, and use only the synthesizability value network during PDVN training. 

% we ignore the cost of reactions and care only about the synthesizability as in most of the previous works. This is implemented by setting the cost value to be $V^{\rm cost}(s) = 0$ for all molecule $s$ while using only the synthesizability value network. We simply name this variant synthesizability.

As shown in Table~\ref{tab:ablation_study}, both variants have lower success rates than PDVN, which implies that 1) it is beneficial to decompose the total cost into dual value networks;
2) cost value network benefits the training, along with synthesizability value network.
% optimizing the cost term can benefit the synthesizability, and it is important to decouple the cost from synthesizability. 
Besides, we observe that PDVN w/o Cost can achieve a slightly higher success rate than SingleValue for the Retro*-0 planner, but the average length of found routes is longer.  
That may suggest that it is important to take 
% SingleValue provides lower-cost routes  
the cost term into consideration.
% SingleValue is able to provide low-cost routes but it suffers from bad synthesizability. In contrast, the second ablation has comparable synthesizability with PDVN, but the route quality is inferior.

% \paragraph{Training on USPTO test dataset.}

\subsection{Qualitative Case study}

\begin{figure}[t]
\resizebox{\columnwidth}{!}{%
\includegraphics[width=0.5\textwidth]{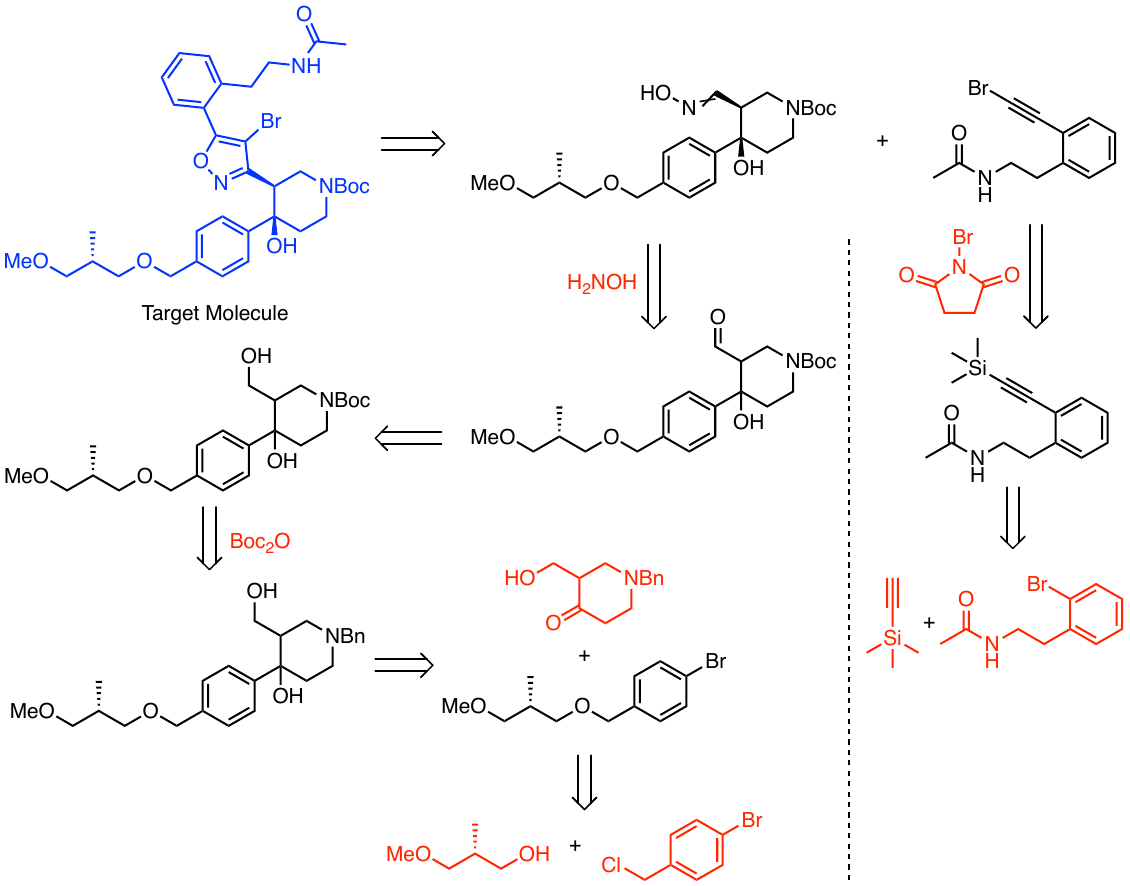}%
}
%\centerline{\includegraphics[width=0.3\textwidth]{case_study.png}}
\caption{
{
Case study of an exemplary route predicted with PDVN. The arrow represents the single-step chemical reaction,
% in the direction of retrosynthesis reaction, 
% and the plus sign means there are multiple reactants. 
and the molecules at the end of the synthesis route are building block molecules.
}
}
\label{fig:case_study}
\end{figure}

A potential risk of reinforcement learning is exploitation of the environment, i.e. the single step SL model with its known imperfections. We performed a qualitative analysis of the routes given by Retro* using different models. The analysis indicated that PVDN leads to routes of similar chemical plausibility as the SL model, with usually fewer steps. 
As an example, we choose molecule 25 from the USPTO 190 test dataset. With the computation budget of 500 model calls, Retro* cannot solve it using the SL model, however, with PDVN it is able to provide a route. 
As shown in Fig.~\ref{fig:case_study}, our method identifies a route closely related to the hold-out original route, albeit with one synthesis step less. 

\section{Conclusion}

% \textbf{Summarize our work again.}
In this work, we introduce PDVN, a novel policy learning framework for retrosynthesis. 
PDVN improves the single-step predictor to not only predict valid reactions, but also predict reactions that lead to synthesizable and low-cost synthesis routes.
Experiments on the widely-used USPTO dataset demonstrate that PDVN significantly enhances both the search success rate and route quality of existing multi-step planners (e.g., Retro*, RetroGraph), achieving state-of-the-art performance on the USPTO dataset.  
For future work, one potential direction is to extend our PDVN algorithm to other single-step models, such as template-free ones, which have shown great single-step accuracy and generalizability.
%Another interesting direction is to increase the amount of training target molecules and test the performance of PDVN on more real-world target molecules. 
We anticipate that our algorithm will help to further accelerate the discovery of molecules and materials for health care, agriculture, and energy storage.

\section{Acknowledgments}
We would like to thank Elise van der Pol for proofreading the manuscript and offering invaluable feedback, as well as Sarah Lewis and Megan Stanley for their insightful discussions. 
Di Xue and Zongzhang Zhang acknowledge funding from the National Science Foundation of China (62276126) and the Fundamental Research Funds for Central Universities (14380010).

% \textbf{Future work: 1. scalability: training on more drugs. 2. extend to template-free single-step models, e.g., Transformers. 3. care about the diversity of the generated routes.}

\nocite{langley00}

\bibliography{example_paper}
\bibliographystyle{icml2023}

%%%%%%%%%%%%%%%%%%%%%%%%%%%%%%%%%%%%%%%%%%%%%%%%%%%%%%%%%%%%%%%%%%%%%%%%%%%%%%%
%%%%%%%%%%%%%%%%%%%%%%%%%%%%%%%%%%%%%%%%%%%%%%%%%%%%%%%%%%%%%%%%%%%%%%%%%%%%%%%
% APPENDIX
%%%%%%%%%%%%%%%%%%%%%%%%%%%%%%%%%%%%%%%%%%%%%%%%%%%%%%%%%%%%%%%%%%%%%%%%%%%%%%%
%%%%%%%%%%%%%%%%%%%%%%%%%%%%%%%%%%%%%%%%%%%%%%%%%%%%%%%%%%%%%%%%%%%%%%%%%%%%%%%
\newpage
\appendix
\onecolumn
% \section{You \emph{can} have an appendix here.}

% You can have as much text here as you want. The main body must be at most $8$ pages long.
% For the final version, one more page can be added.
% If you want, you can use an appendix like this one, even using the one-column format.

\section{Implementation Details}

\subsection{Network architecture}

The goal of the PDVN training is to optimize the parameters of the policy network and dual value networks. 
The inputs of these networks are binary strings of length $2048$, which are the Morgan fingerprints of molecules of radius $2$.
The policy network has two sub-networks inside it, i.e., the reference single-step model and the learnable single-step model. The two sub-networks share the same MLP structure, and they are both initialized with the parameters from the SL trained model by \cite{chen20retrostar}. The dual value networks also share the same MLP structure but the output activation functions are different. The hyper-parameters of these neural networks are listed below.

\begin{table}[h]
\centering
\caption{Hyper-parameters of neural networks.}
\label{tab:hyper}
\begin{tabular}{cc}
\toprule
Single-step model input units & $2048$ \\
Dual value networks input units & $2048$ \\
Single-step model hidden units & $512$ \\
Dual value networks hidden units & $512$ \\
Cost value network output activation & softplus \\
Synthesizability value network output activation & sigmoid \\
\bottomrule
\end{tabular}
\end{table}

\subsection{PDVN planning phase}

To generate experiences for training, we design an MCTS-based planner based on dual value networks. Our planner resembles the online MCTS planner that conducts a fixed number of simulations at the root node in each iteration. We use a queue to store the simulation roots.
Within each simulation, we alternately select reaction and molecule nodes until a leaf molecule node is encountered. 
We use the PUCT rule to select a reaction (in Eqn.~\ref{eq:puct}), where a parameter $C$ balances the trade-off between exploitation and exploration.
Besides, we set a maximum route depth to avoid too-long synthesis routes. 
% in favor of short synthesis routes. 
To avoid circular routes, we further eliminate those reactions that appeared in their ancestors.

\begin{table}[h]
\centering
\caption{Hyper-parameters for PDVN planning.}
\label{tab:hyper}
\begin{tabular}{cc}
\toprule
C (PUCT) & 1.0 \\
$\alpha$ (Synthesizability penalty) & 0.8 \\
MCTS depth & 15 \\
Number of simulations & 100 \\
$c_{\rm dead}$ & 5.0 \\
$c_{\rm rxn}(s, a)$ & 0.1 \\
\bottomrule
\end{tabular}
\end{table}

\subsection{PDVN training}

The training of the PDVN algorithm iterates over the whole training target molecule dataset $\mathcal{D}_{\rm train}$ for three epochs. For each update, we uniformly sample a batch of training target molecules from $\mathcal{D}_{\rm train}$ to generate the training data, and update the networks. To speed up the data generation process, we implement a parallel version of MCTS planners to run MCTS planning for multiple training target molecules simultaneously. 
% When the searches are done, we update three neural networks based on the generated dataset.
% We save one checkpoint every $25000$ target molecules have been trained.
The whole training process takes about $18$ hours on a server with four NVIDIA TITAN Xp and 48 CPU cores
(using $15$ parallel workers).

\begin{table}[h]
\centering
\caption{Hyper-parameters for PDVN training.}
\label{tab:hyper}
\begin{tabular}{cc}
\toprule
Training dataset size & 299202 \\
Batch size & 1024 \\
Optimizer & Adam \\
Learning rate & 1e-3 \\
Dropout rate & $0.1$ \\
Mini-batch size & $128$ \\
SL epochs & 8 \\
\bottomrule
\end{tabular}
\vskip-0.11in
\end{table}

\end{document}